\documentclass{article}

\usepackage{microtype}
\usepackage{graphicx}

\usepackage{booktabs}
\usepackage{hyperref}

\usepackage{times}
\usepackage{color,graphicx,amsmath,amssymb,epsfig,epsf,xspace,setspace,dsfont}
\usepackage[small,bf]{caption}
\usepackage{multicol, multirow}
\usepackage[algo2e,linesnumbered,ruled]{algorithm2e} 
\usepackage{graphicx}
\usepackage{soul}
\usepackage{colortbl}
\usepackage{rotating}
\usepackage{times}
\usepackage{textcomp}
\usepackage{url}
\usepackage{epsf}
\usepackage{wasysym}
\usepackage{mdwtab}

\usepackage{bm}
\usepackage{mdframed}

\usepackage{subcaption}
\newmdtheoremenv{theo}{Theorem}

\def \R{\mathbb R}

\newtheorem{lemma}{Lemma}
\newtheorem{definition}{Definition}
\newtheorem{proof}{Proof}
\def \cS{{\cal S}}
\def \cJ{{ \cal T}}
\def \cT{{ \cal R}}

\def \xb{\bar{\beta}}

\def \d1{\mathds{1}}

\def \wb{\widehat{\beta}}
\def \tb{\tilde{\beta}}

\newcommand\tab[1][0.2cm]{\hspace*{#1}}

\def\R{{\mathbb R}}

\title{\LARGE \bf
Scalable Sparse Subspace Clustering\\ via Ordered Weighted $\ell_1$ Regression
}

\author{Urvashi Oswal$^{1}$ and Robert Nowak$^{2}$
\thanks{$^{1}$Urvashi Oswal is with the Department of Electrical Engineering, University of Wisconsin-Madison, Madison, WI 53706, USA
        {\tt\small uoswal@wisc.edu}}%
\thanks{$^{2}$Robert D. Nowak is with the Department of Electrical Engineering, University of Wisconsin-Madison, Madison, WI 53706, USA
        {\tt\small rdnowak@wisc.edu}}%
}

\begin{document}

\date{}

\maketitle
\thispagestyle{empty}
\pagestyle{empty}

\begin{abstract}
  The main contribution of the paper is a new approach to subspace
  clustering that is significantly more computationally efficient and
  scalable than existing state-of-the-art methods. The central idea is
  to modify the regression technique in sparse subspace clustering
  (SSC) by replacing the $\ell_1$ minimization with a generalization
  called Ordered Weighted $\ell_1$ (OWL) minimization which performs
  simultaneous regression and clustering of correlated variables.
  Using random geometric graph theory, we prove that OWL regression
  selects more points within each subspace, resulting in better
  clustering results.  This allows for accurate subspace clustering
  based on regression solutions for only a small subset of the total
  dataset, significantly reducing the computational complexity
  compared to SSC. In experiments, we find that our OWL approach
  can achieve a speedup of 20$\times$ to 30$\times$ for synthetic problems and 4$\times$ to 8$\times$ on real data problems.
\end{abstract}

\section{Introduction}\label{sec:intro}
Subspace clustering refers to the task of grouping high-dimensional data points into distinct subspaces.  This generalizes classical, single-subspace approaches to data modeling like principal components analysis (PCA).  Effectively, subspace clustering aims to represent data in terms of a union of subspace (UoS).  Many applications, ranging from computer vision  (\textit{e.g.}, image segmentation \cite{cv1}, motion segmentation \cite{cv2} and face clustering \cite{cv3}) to network analysis \cite{network}, have demonstrated the advantages of this generalization.  Unlike classical PCA, the problem of fitting a UoS model to data is a computationally challenging task, and numerous approaches have been proposed. For a comprehensive review of these algorithms, we refer the reader to the tutorial \cite{tutorial}. The state-of-the-art is Sparse Subspace Clustering (SSC) \cite{ssc} which provides both tractability and provable guarantees under mild conditions \cite{geometric}. 

SSC is a computationally intensive method.  It requires performing a sparse regression for each of the $N$ points in the dataset of interest. The main contribution of this paper is a new approach that has the potential to significantly reduce the computational complexity, making it more applicable to large-scale problems.   The central idea is to modify the regression technique so that accurate clustering is possible using only the results of a $k\ll N$ regressions, instead of all $N$.  This reduces the complexity by a factor of $N/k$. The modified regression is based on the Ordered Weighted $\ell_1$ (OWL) regularizer, which performs simultaneous regression and clustering of correlated variables.  The clustering property of the OWL, combined with ideas from random geometric graph theory, allows us to prove that the new approach, called OWL Subspace Clustering (OSC), tends to select more points from the correct subspaces in each regression compared to SSC.    In the ideal case, where $L$ subspaces are orthogonal and 
the number of points per subspace is sufficiently large, then OSC can succeed with just $L \ll N$ optimizations (gain factor of roughly $N/L$) as detailed later in Section~\ref{sec.orthogonal}. This key feature of OSC makes accurate clustering possible based on regression solutions for only a small subset of the total
  dataset, significantly reducing the computational complexity
  compared to SSC. In experiments, we find that OSC
  can achieve a speedup of 20$\times$ to 30$\times$ even for small
  scale synthetic problems.

\subsection{Relation to prior work}
\begin{figure}[t!]
\centering
\includegraphics[width=0.5\textwidth]{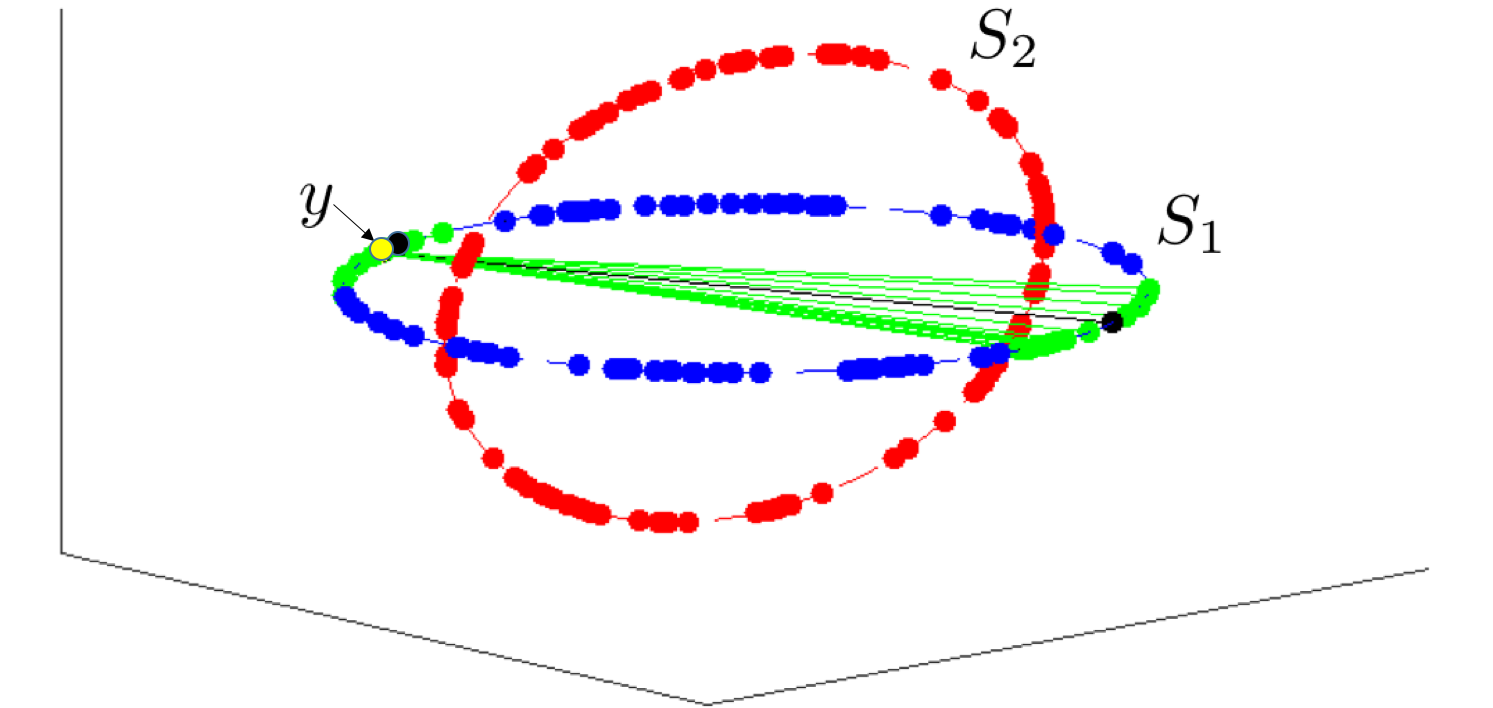}
\caption{2-dimensional subspaces in $\R^3$, $\cS_1$ and $\cS_2$. The points with non-zero coefficients in solutions of $\ell_1$ and Ordered Weighted $\ell_1$ sparse regressions on $y$ (yellow point) depicted as black and green points (and edges) respectively. The automatic clustering property of OWL leads to a hybrid approach that tends to select same points as SSC and their Euclidean distance based neighbors.}
\label{fig:hybrid}
\end{figure}

When the data is high-dimensional or the number of data points is large, solving the $N$ Lasso problems (each in $N - 1$ variables) in SSC can be computationally challenging. Greedy algorithms for computing sparse representations of the data points (in terms of all the other data points) have therefore been popular alternatives. Broadly speaking, three kinds of such algorithms have been proposed in the literature, namely Thresholded subspace clustering \cite{heckel2015robust}, subspace clustering using Orthogonal Matching Pursuit (OMP) \cite{dyer2013greedy, you2016scalable}, and Dimensionality-reduced subspace clustering \cite{wang2015deterministic, heckel2017dimensionality}. TSC relies on the nearest neighbors--in spherical distance--of each data point to construct the adjacency matrix. SSC-OMP employs OMP instead of the Lasso to compute sparse representations of the data points. Similar in spirit to SSC-OMP, Nearest subspace neighbor (NSN) \cite{park2014greedy} greedily assigns to each data point a subset of the other data points by iteratively selecting the data point closest (in Euclidean distance) to the subspace spanned by the previously selected data points. In this paper, we propose OWL Subspace Clustering (OSC), which employs the Ordered Weighted $\ell_1$ (OWL) regularizer instead of the Lasso to compute sparse representations of the data points. The clustering property of OWL (discussed next) leads to a hybrid approach between SSC and spherical distance based methods like TSC (illustrated in Figure \ref{fig:hybrid}). The effect is that OSC tends to assign non-zero weights to the same points as SSC and additionally selects neighbors of these points in terms of Euclidean distance. All of the above methods, including OSC, can be carried out on dimensionality reduced data points, in the spirit of \cite{wang2015deterministic, heckel2017dimensionality}. We compare the performance of these methods in Section \ref{sec:realexp}. 

\subsection{Review of SSC}

The key idea of SSC is representing each data point as a sparse linear combination of the remaining data points. The rationale is that points in the same subspace are likely to be selected to represent a given point, and thus the selected points provide an indication of the cluster. Sparse linear regression using $\ell_1$ minimization is used to determine the representations for each point. SSC solves $N$ sparse linear regressions over $N$ data points to form an $N \times N $ adjacency matrix, which in turn defines a graph where the vertices are data points with edges indicated by the adjacencies. Spectral clustering is used to partition the graph, and hence the data, into clusters. Each sparse linear regression solves 
\begin{equation}\label{eqn:l1opt}
\min_{\beta \in \R^N} \|\beta\|_1 \textnormal{ such that } y = X\beta,
\end{equation}
where the columns of $X \in \R^{n \times N}$ represent the $N$ $n$-dimensional data points and $y \in \R^n$ is a data point that is to be represented as a linear combination of the columns using the sparse coefficient vector $\beta$. Another sparse regression commonly used in practice is the Lasso given by 
\begin{equation}\label{eqn:lasso}
\min_{\beta \in \R^N} \frac{1}{2}\|y - X\beta\|_2^2 + \lambda\|\beta\|_1,
\end{equation}
for $\lambda > 0$. It is sometimes referred to as the Lagrangian or ``noisy" version since it is used for subspace clustering with noise \cite{noisy, soltanolkotabi2014robust}.

\subsection{OWL norm and its clustering property}
\begin{figure*}[t!]
\centering
\includegraphics[width=0.99\textwidth]{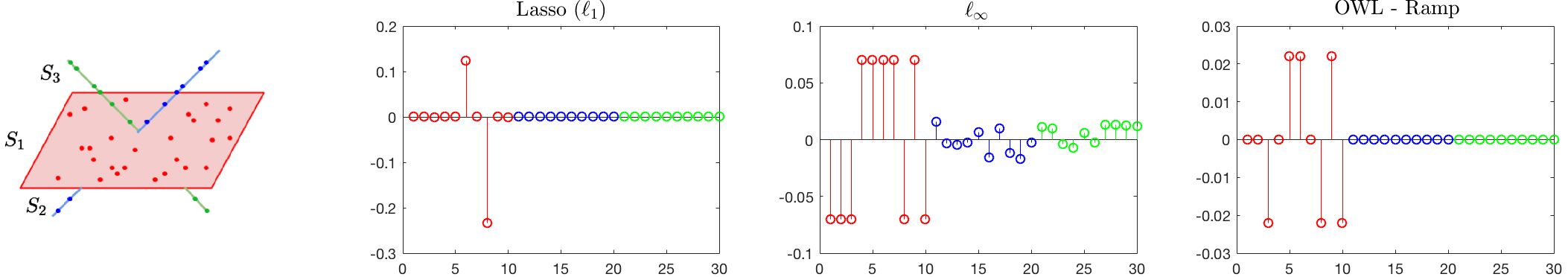}
\caption{Solutions of the $\ell_1$, $\ell_{\infty}$, and OWL minimizations for $y_i$ lying in $\cS_1$. 10 points selected from each of the three subspaces ordered such that the first and the last 10 points belong to $\cS_1$ and $\cS_3$ respectively. The $\ell_1$ solution corresponds to choosing two other points lying in $\cS_1$ whereas the $\ell_{\infty}$ solution selects points from all subspaces. OWL selects more points from $\cS_1$ than $\ell_1$.}
\label{fig:sscexamples}
\end{figure*}

A new family of regularizers, called Ordered weighted $\ell_1$ (OWL), for sparse linear regression was recently introduced and studied in \cite{zeng,slope,aistatsOWL,oscar}. The authors show that OWL automatically clusters and averages regression coefficients associated with strongly correlated variables. This has a desirable effect of selecting more of the relevant variables than the $\ell_1$ penalty. The main idea in this paper is to leverage this feature of OWL for subspace clustering. OWL tends to select more points from the common subspace in each regression, compared to $\ell_1$ methods (see Figure \ref{fig:sscexamples}), which consequently improves the performance of spectral clustering.

\begin{figure}[t!]
\centering
\includegraphics[width=0.75\textwidth]{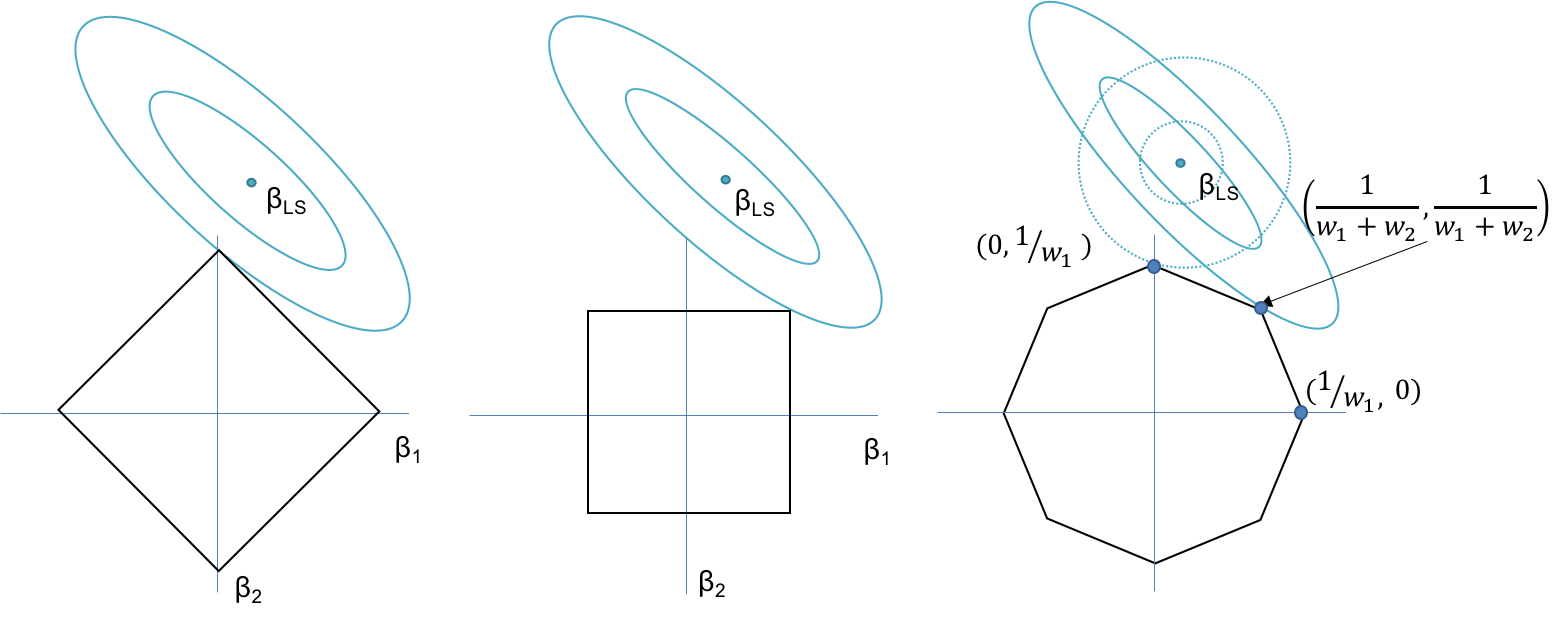}
\caption{OWL balls in $\R^2$ with different choices of weights leading to the $\ell_1$ and $\ell_{\infty} $ as special cases. (Left) $w_1 = w_2 > 0 $ and $\Delta_w = 0$. (Middle) $w_1 > w_2 = 0$ and $\Delta_w = w_1$.  (Right) $w_1 > w_2 > 0$. Contours centered at Least squares estimate ($\beta_{LS}$). \\ Low correlation (dotted), leads to solution at $\beta_1 = 0$. \\High correlation (solid), leads to clustered solution at $\beta_1 = \beta_2$.}
\label{fig:owlballs}
\end{figure}

The OWL norm \cite{zeng, slope} is defined as
\begin{equation} \label{eqn:owldef}
\Omega_w(\beta) = \sum_{i=1}^N w_i |\beta|_{[i]}, 
 \end{equation}
where $|\beta|_{[i]}$ denotes the $i$-th largest magnitude,  $w \in \R^N_+$ is a vector of weights, such that $w_1 \geq \dots \geq w_N \geq 0$ and $w_1 > 0$. The OWL norm reduces to the $\ell_1$ norm when all the weights are set to be equal and the $\ell_{\infty}$ norm when $w_2 = \dots = w_N = 0$. It is also useful to look at the structure of the OWL norm balls. Figure \ref{fig:owlballs} shows the OWL balls in $\R^2$. 

The OWL-regularized regression used in this paper is 
\begin{equation}\label{eqn:owlopt}
\min_{\beta \in \R^N} \frac{1}{2} \|y-X\beta\|_2^2 + \Omega_w(\beta).
\end{equation}

The following result provides a sufficient condition stating that when columns of $X$ are correlated enough, OWL automatically clusters them (i.e., correlated columns will have equal-valued coefficients in the solution). Let  $\Delta_w = \min\{w_l- w_{l+1}, l = 1, ..., N-1\}$ be the minimum gap between consecutive elements of $w$. 
\begin{lemma}[Theorem 2.1 in \cite{aistatsOWL}]\label{thm:owl}\emph{
Let $\wb$ be a solution of (\ref{eqn:owlopt}), and $x_i$ and $x_j$ be two columns of $X$. If $\|x_i - x_j \|_2 < \Delta_w/\|y\|_2$, then $|\wb_i| = |\wb_j|$.}
\end{lemma}
In this paper, we propose to replace the $\ell_1$ minimization in SSC with the OWL regularizer in order to discover many more points from the true subspace in each optimization. The key observation is that points in the same subspace will be more correlated (in the sense above), than points in different subspaces.  We formalize this idea using tools from random geometric graph theory. Next, we demonstrate the benefit of the clustering property of OWL in a simple setting where the subspaces are orthogonal to each other.

\subsection{Orthogonal subspaces example}
\label{sec.orthogonal}
To build some intuition, let $X \in \R^{n \times N}$ be a matrix whose columns are drawn from a union of $L$ \textit{orthogonal} linear subspaces, $\cS_{1} \cup \dots \cup \cS_L$. Let $y$ be a new point from subspace ${\cS}_{\ell}$.  To keep the notation simple, let us assume that the dimension of each subspace is $d$. Suppose $\cJ$ contains the indices of columns belonging to subspace $\cS_{\ell}$ and $|\cJ| = N_{\ell}$. 

It is easy to see that any solution of (\ref{eqn:owlopt}), $\wb$, satisfies $\wb_j = 0$ for all $j \notin \cJ$, since columns from orthogonal subspaces cannot reduce the residual error and including them will increase the OWL penalty.  Additionally, assume that points in the subspace are uniformly drawn from the unit hypersphere in that subspace. Then consider the graph constructed by placing an edge between pairs of points (vertices) that are within $\Delta_w$ Euclidean distance of each other (where $\Delta_w$ is the minimum gap in the OWL weights). A simple argument (developed in Section \ref{thm2sketch}) shows that the resulting random geometric graph is connected with probability at least $1-\delta$, if the number of points in the subspace is large enough, specifically $N_{\ell} \propto \Delta_w^{-d} \log(\Delta_w^{-d}/\delta)$.  It then follows from Lemma~\ref{thm:owl} that $|\wb_j| = |\wb_i|$ for $\forall i,j \in \cJ$. In other words, all the columns within the subspace will be selected and have equal-valued coefficients.

 \begin{figure}[t!]
\centering
\includegraphics[width=0.75\textwidth]{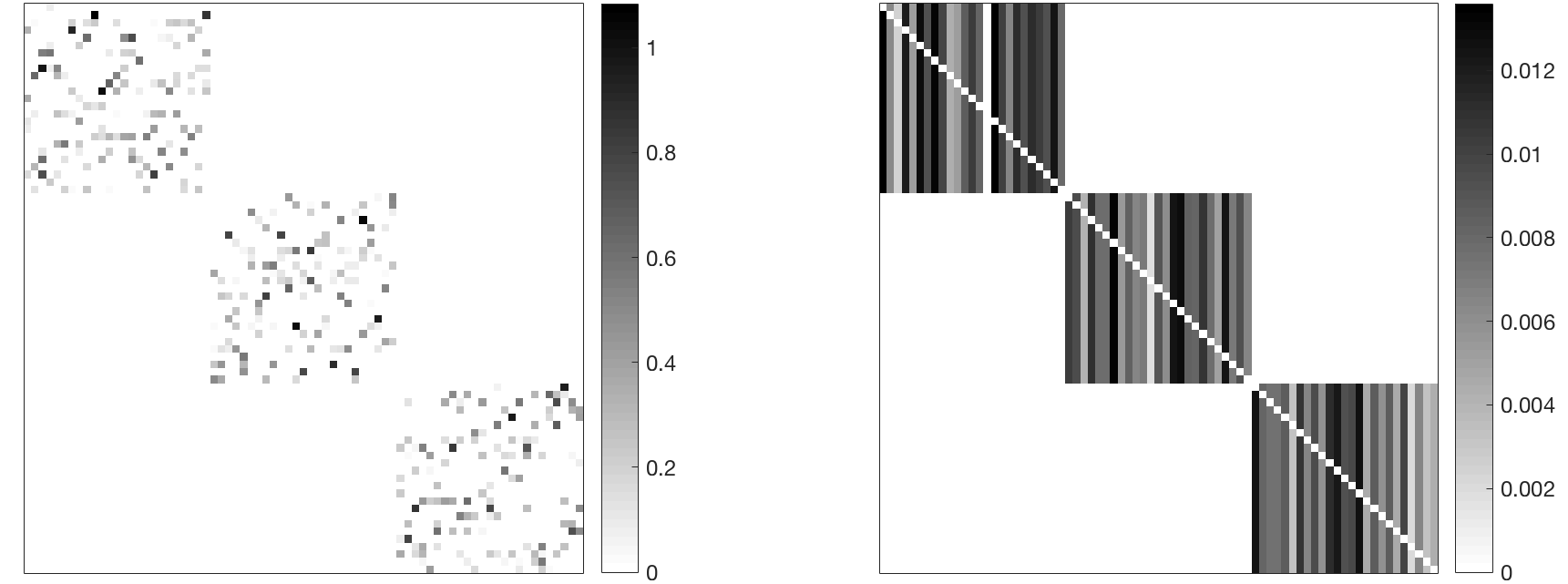}
\caption{Examples of coefficient matrices $|B| = [|\wb_1| \dots |\wb_N|]$ for exact $\ell_1$ minimizations (left) and OWL optimizations (right) with the contiguous columns lying in three orthogonal subspaces each of dimension $d = 5$ in $\R^{15}$. The plots were generated using OWL-Ramp weights defined in Section \ref{sec:theory}. }
\label{fig:orthogonal}
\end{figure}

Figure \ref{fig:orthogonal} shows an example of the coefficients generated by performing the optimizations in (\ref{eqn:l1opt}) and (\ref{eqn:owlopt}) on a collection of points from three orthogonal subspaces. The results obtained with orthogonal subspaces suggest that, in an ideal scenario, we could perform subspace clustering by running only \textbf{one} OWL optimization per subspace without the need for spectral clustering leading to a total of only $L$ optimizations where $L$ is the number of subspaces. This can easily be achieved by running one OWL optimization on a randomly selected point, removing the points chosen by OWL and repeating the process till no points remain. This works since \textit{OWL has the potential to find all the points in the same subspace} in a single run. This has the potential of significantly reducing the computational complexity of SSC. However, OWL could lead to clustering of points from other subspaces when subspaces are not orthogonal. This tradeoff is studied in more detail in the rest of the paper by providing theoretical and empirical evidence suggesting that an OWL-based fast subspace clustering algorithm (stated in Section \ref{sec:algo}) can reduce the computational complexity of SSC by a factor depending on the nature of the subspaces.

\section{Algorithm}\label{sec:algo}
The SSC procedure can easily be modified by replacing the sparse linear regression step with the OWL optimization in (\ref{eqn:owlopt}). As seen in the example with orthogonal subspaces, OWL solutions are denser due to its clustering property. This leads to an intuitive extension of the algorithm where we run only a subset of the total $N$ possible optimizations. By running a random subset of $k$ optimizations the computational complexity can be reduced. We formalize this version of OSC in Algorithm \ref{owlalgo}. 

\begin{algorithm2e}[t]

\caption{OSC with $k$ random seeds}\label{owlalgo}

\SetKwInOut{Input}{Input}
    \SetKwInOut{Output}{Output}

    \Input{A set of data points $X \in \R^{n\times N}$, $k \in \{1,\dots ,N\}$.}
Initialize $B = \boldsymbol{0}_{N \times N}$.\\
For $i \in \{1,\dots ,k\}$,\\
\tab Randomly select an index $j_i$ from $[N]$. \\
\tab Obtain $\wb$ by regressing $y = X_{\cdot j_i}$ onto the remaining columns of $X$ using the OWL minimization (\ref{eqn:owlopt}). \\
\tab Store $B_{\cdot j_i} = \wb$ with $B_{j_i, j_i} = 0$.\\
Form affinity matrix $W = |B| + |B|^T$. \\
Apply spectral clustering to the Laplacian of $W$ to obtain a partition.

\Output{Subspaces $\{\cS_{\ell}\}^L_1$, Cluster labels.}

\end{algorithm2e}

The intuition developed in the case of orthogonal subspaces suggests that taking $k = L$, where $L$ is the number of subspaces may suffice for the OWL approach.  Thus, the proposed algorithm requires solving $O(L)$ sparse regressions, whereas traditional SSC involves solving $O(N)$, and typically $L\ll N$.  It is also worth noting that the computational complexity of lasso and OWL is essentially the same.  Obviously, one could also consider reducing the computational complexity of traditional SSC by also using only $O(L)$ regressions instead of all $N$.  However, as shown in our experiment section, SSC performance quickly degrades when the number of regressions is reduced, while the proposed OWL-based algorithm's performance does not, leading to a speedup of up to 30$\times$ in some cases.

\section{Theoretical results}\label{sec:theory}
In this section, we begin by formalizing the notations and model, followed by the main results analyzing the behavior of the solution of the optimization (\ref{eqn:owlopt}). We also provide examples and remarks to understand the theoretical results along with proof sketches. The proof sketches also provide the intermediate results ranging from sparse regression with correlated variables to random geometric graph theory used to prove the main results.

\subsection{Notation and model}
We are given data points lying in a union of unknown linear subspaces; there are $L$ subspaces $\cS_1, \cS_2, \dots, \cS_L$ of $\R^n$ of dimensions $d_1, d_2, \dots, d_L$.  We are given a collection of $N$ data points as columns of $X \subset \R^{n\times N}$, which may be partitioned as $X =[ X_1,  X_2, \dots X_L]$ without loss of generality; for each $\ell \in \{1, \dots, L\}, X_{\ell}$ is a collection of $N_{\ell}$ vectors that belong to subspace $\cS_{\ell}$. The goal is to approximate the underlying subspaces using the points in $X$. We also assume the columns are normalized to have unit norm, $\|x_i\|_2 = 1$. The notation used is summarized in Table~\ref{table:notation}.

\begin{table}[h!]
\centering
\caption{Notation and parameters}
\begin{tabular}{ | c | l | }
\hline
$ L$ & Number of subspaces  \\ 
$d_{\ell}$ & Dimension of each subspace for $\ell = 1\dots L$  \\  
$N_{\ell}$ & Number of points sampled from each subspace \\    
$\rho_{\ell}$ & Sampling density $\rho_{\ell} = N_{\ell}/d_{\ell}$\\
$N$ & Total number of points, $ N = \sum_{\ell} N_{\ell}$\\
$n$ & Ambient dimension\\
$\lambda$ & $\ell_1$ component of OWL-Ramp, $\lambda > 0$\\
$\Delta$ & Slope of OWL-Ramp, $\Delta \geq 0$\\
$r$ & Length of OWL-Ramp, $1 \leq r \leq N$\\
$\alpha$ & Normalized maximum affinity between subspaces, \\
& $0 \leq \alpha \leq 1$ \\
$k$ & Number of optimizations $1\leq k \leq N$\\
$\sigma$ & Noise level\\
\hline
\end{tabular}
\label{table:notation}
\end{table}

We consider the intuitive \textit{semi-random model} introduced in \cite{geometric} where the subspaces are fixed, and points are distributed randomly on each of the subspaces. To measure the notion of closeness or correlation between two subspaces, the affinity between subspaces is used.

\begin{definition} \emph{The principal angles $\{\theta^{(i)}\}_{i = 1}^{(d \wedge d')}$ between subspaces $\cS$ and $\cS'$œ of dimensions $d$ and $d'œ$, are defined by
$$\cos(\theta^{(i)}) = \max_{u \in \cS} \max_{v \in \cS'} \frac{ u^T v }{\|u\| \| v\|}  := \frac{u^T_i v_i}{\|u_i\| \|v_i\|}$$
with orthogonality constraints $u^Tu_j = v^T v_j = 0, j = 1,\dots, i -1$.}
\end{definition}

\begin{definition}\emph{The normalized affinity between subspaces is
$$ \textnormal{aff}(\cS,\cS') = \sqrt{\frac{\cos^2(\theta^{(1)}) + \dots + \cos^2(\theta^{(d \wedge d')})}{d \wedge d'}} $$}
\end{definition}
The affinity is low when the subspaces are nearly orthogonal and high when the subspaces overlap significantly (it is equal to one when one subspace is contained in the other). Hence, when the affinity is high, clustering is hard whereas it becomes easier as the affinity decreases.

\subsection{Performance metrics}
To quantify performance of the algorithm, we use the following metrics from \cite{soltanolkotabi2014robust}

\textit{False discovery:} Fix $i$ and $j \in \{1, \dots, N\}$ and let $B$ be the outcome of Step1 in Algorithm 1. Then we say that $(i, j)$ obeying $B_{ij} \neq 0$ is a false discovery if $x_i$ and $x_j$ do not belong to the same subspace.

\textit{True discovery:} Similarly, we say that $(i, j)$ obeying $B_{ij} \neq 0$ is a true discovery if $x_i$ and $x_j$ belong to same subspace.

\subsection{Main results}\label{sec:mainresults}
We state the first main result showing that the solution of the OWL optimization does not include false discoveries if the the affinity between the subspaces is small enough, \textit{i.e.,} the subspaces are not too close. 

Let $\bar{w}_{N_{\ell}+1} := \frac{1}{N-N_{\ell}-1}\sum_{j=N_{\ell}+1}^{N} w_j$ be average of the tail of the OWL weights used.
\begin{theo}\label{thm:nofalsedisc}\emph{
If $\cS_{\ell}$, the subspace to which the $i$-th column belongs, obeys
\begin{equation}\label{eqn:upper}
\alpha_{\ell} := \max_{k: k \neq \ell} \ \textnormal{aff}(\cS_{\ell}, \cS_k)\  \leq \   \kappa_0 \frac{\bar{w}_{N_{\ell} + 1}}{w_1} \frac{\sqrt{\log N_{\ell}/d_{\ell}}}{\log N}
\end{equation}
where $\kappa_0$ is a fixed numerical constant, then $\wb_j = 0$ for all $x_j \notin \cS_{\ell}$, \textit{i.e.,} there is no false discovery in the $i$-th column of $B$ with probability at least $1-L(4/N^2+e^{-\sqrt{N_{\ell} d_{\ell}}})$.}
\end{theo}
Roughly stated the result says that with high probability the OWL solution contains no false discoveries if the ratio of OWL parameters is big enough for fixed subspaces. The affinity is higher for overlapping subspaces, then we must reduce the gaps in the OWL weights ($\Delta_w$ in the introduction).  This is roughly equivalent to making the ratio of weights nearly 1. As seen in the introduction, the clustering property of OWL depends on gaps in consecutive weights. OWL clusters columns that are at most the gap away from each other. Hence, making the weight ratio close to 1 results in reducing the weight gaps and in turn reduces the radius of OWL clustering. For smaller values of affinity, the OWL gap can be made bigger and OWL will then group (assign equal coefficients to) more points. Thus, there is a tension in the closeness of the subspaces and the radius or gap of OWL clusters. For higher values of affinity, OWL with bigger weight gaps does not provide any benefit over Lasso hence we must set the weight ratio to one. The weight ratio is exactly one for the $\ell_1$ penalty and our bound for the special case of Lasso matches \cite{geometric} for SSC. 

\begin{figure}[t!]
\centering
\includegraphics[width=0.5\textwidth]{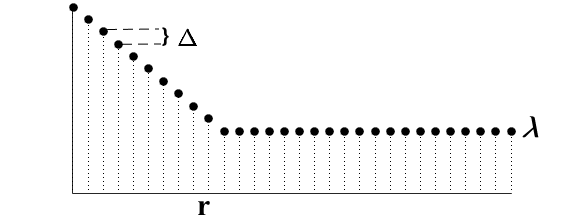}
\caption{An example of OWL-Ramp weights.}
\label{fig:ramp}
\end{figure}

As foreshadowed in the introduction, we show that all the columns within the subspace of interest will have equal-valued coefficients if enough points are sampled, specifically $N_{\ell} \propto 1/\Delta_w^{d}$. Recall that $\Delta_w$ is the minimum gap in the OWL weights. Usually, this gap is very small, like O($1/N$) or in some cases zero, which makes the requirement infeasible. Hence, as part of the second main result, we will prove new clustering bounds for a specific type of the OWL norm called OWL-Ramp with weights 
 \begin{align*}
 w_i &= (r-i+1)\Delta + \lambda &\textnormal{ for } i \in \{1,\dots,r\},\\
 & = \lambda &\textnormal{ for } i \in \{r+1, \dots, N\},
 \end{align*}
where $\lambda > 0$, $\Delta \geq 0$ and $1\leq r \leq N$. Figure \ref{fig:ramp} depicts an example of the OWL-Ramp weights. The OSCAR regularizer \cite{oscar} is a special case obtained by setting $r = N$. Note that if $r \leq N_{\ell}$, then $\bar{w}_{N_{\ell}+1} = \lambda$ and the affinity condition in Theorem \ref{thm:nofalsedisc} becomes
 $$\alpha_{\ell} \  \leq \   \kappa_0 \frac{\lambda}{\lambda+r\Delta} \frac{\sqrt{\log \rho_{\ell}}}{\log N}$$

\begin{theo}\label{thm:truedisc}\emph{
Let $\wb$ be a solution of (\ref{eqn:owlopt}) with OWL-Ramp weights. If $r \leq N_{\ell}$, the conditions in Theorem \ref{thm:nofalsedisc} are satisfied, and the
\begin{equation}\label{eqn:lower}
N_{\ell} > \kappa_1 \Delta^{-d_{\ell}} \log(\Delta^{-d_{\ell}}/\delta)
\end{equation}
points within the subspace of interest are sampled uniformly at random from the unit hypersphere, then the set $M = \{j: |\wb_j| =  \max_i |\wb_i|\}$ has cardinality greater than or equal to the ramp parameter, $r$, with probability at least $1- \delta - L(4/N^2+e^{-\sqrt{N_{\ell} d_{\ell}}})$. ($\kappa_1$ is a fixed numeric constant.)}
\end{theo}
The result roughly says that if enough points are sampled from the corresponding subspace then with high probability the top $r$ coefficients in the OWL solution have equal magnitude. Combining this with Theorem \ref{thm:nofalsedisc}, if the solution to OWL optimization is non-trivial, this is equivalent to making $r$ true discoveries.

It is easy to see that $\wb = \b0$,  if $\Omega^*_w(X^Ty) < 1$ where $\Omega^*_w(\beta)$ is the dual norm of OWL defined later in the section. In order to ensure that the solution is non-trivial we need at least $\bar{w} \leq \|X^Ty\|_{\infty}$ for $\bar{w} = \sum_{j = 1}^N w_j /N $. The $\|X^Ty\|_{\infty}$ term scales at most like $\sqrt{(\log N)/d}$ and for OWL-Ramp, $\bar{w} \approx \lambda$. Intuitively, the $\ell_1$ component needs to be made small enough to achieve a non-trivial solution. 

To make sense of the results, the condition in (\ref{eqn:upper}) can be rewritten as an upper bound on $\Delta$ and the condition in (\ref{eqn:lower}) as a lower bound on $\Delta$. Ignoring constants and $\log$ terms
\begin{equation*}\Delta \lesssim \frac{\lambda}{r} \left( \frac{1}{\alpha_{\ell}} - 1\right)\textnormal{ and }\Delta \gtrsim \left( \frac{1}{N_{\ell}} \right)^{1/d_{\ell}}
\end{equation*}
Intuitively, we want to make $\Delta$ small enough so that nothing from outside of the true subspace is selected, but at the same time we want to make $\Delta$ as big as possible so many points are selected from the true subspace. This leads to a trade-off between the number of false discoveries and true discoveries. For orthogonal subspaces or $\alpha_{\ell} = 0$, the upper bound is trivially satisfied and we can make $\Delta$ big enough to group all the points in the same subspace as seen in the introduction. For $\alpha_{\ell} \approx 1$ or when the subspace is contained in another subspace, $\Delta$ needs to be set to zero to ensure no false discoveries. Equivalently, the best we can do in this situation is the Lasso solution. Guided by the theory we state rules of thumb for selecting the parameters of OSC followed in the experiments of this paper. 

\subsection{Choice of hyper-parameters.}\label{sec.tuningparameters} 

The $\lambda$ and $\Delta$ parameters are varied in Section~\ref{sec:exp} to demonstrate the trade off between the number of false discoveries and true discoveries. In some applications like motion segmentation, the dimensions of the subspaces are equal and known in advance or can be roughly estimated. In such cases we recommend setting the tuning parameters as follows. 

\begin{itemize}
\item\textbf{Ramp length}, $r$: In general, we recommend setting $r$ proportional to an estimate of number points per cluster since this parameter is related to the number of points clustered by OWL.

\item\textbf{$\ell_1$ component}, $\lambda$:  Informed by theory and experiments in this paper and from SSC literature\cite{soltanolkotabi2014robust}, we set $\lambda$ proportional to $1/\sqrt{d}$. 

\item\textbf{Ramp slope or gap}, $\Delta$: Since the gap in the OWL weights control the clustering behavior the theory suggests setting $\Delta \approx N_i^{-1/d}$ for many true discoveries.  As the affinity of the subspaces increases $\Delta$ needs to be reduced in order to avoid false discoveries. In the experiments, we observe that even small values such as $\Delta = 0.01$ lead to enough clustering in OWL solutions given that there are enough points in each subspace.
\end{itemize}

In the rest of the section, we state the important intermediate results used to prove the main results and provide a sketch of the proofs. Proof details are provided in the Appendix. The intermediate results are stated in generality since they may be useful in other settings. 

\subsection{Proof of Theorem \ref{thm:nofalsedisc}}

We start by proving a deterministic lemma that introduces the OWL dual feasibility condition. First we define the dual norm of the OWL norm given by \cite{zeng}
$$\Omega^*_w(\beta) = \max\{\tau_i\|\beta_{(i)}\|_1, i = 1, \dots, N \}$$ where $\beta_{(i)}$ is the sub-vector of $\beta$,  consisting of the $i$ largest magnitude elements of $\beta$ and $\tau_i = (\sum_{j = 1}^i w_j)^{-1}$. 

\begin{lemma}\label{thm:dual}\emph{
Fix $X \in \R^{n \times N}$ and $\cJ \subset \{1, \dots, N\}$. Suppose $\beta^*$ is a solution to 
$$ \min_{\beta} \frac{1}{2} \|y - X\beta\|_2^2 + \Omega_w(\beta) \textnormal{ subject to } \beta_{\cJ^c} = 0 $$
obeying $\Omega^*_{w'}(X_{\cJ^c}^T(y - X\beta^*)) < 1$ where $w' = [ w_{|\cJ| + 1}, \dots, w_N]$. Then any optimal solution $\wb$ to 
$$ \min_{\beta} \frac{1}{2} \|y - X\beta\|_2^2 + \Omega_w(\beta) $$
must also satisfy $\hat{\beta}_{\cJ^c} = 0$.}
\end{lemma}
Proof details are provided in the Appendix. The lemma tells us that if the OWL dual feasibility condition $\Omega^*_{w'}(X_{\cJ^c}^T(y - X\beta^*)) < 1$ is satisfied for $\cJ = \{j: X_j \in \cS_{\ell}\}$, then we have no false discoveries. To prove Theorem \ref{thm:nofalsedisc}, it suffices to show that the dual feasibility condition is satisfied. One sufficient condition to satisfy the OWL dual feasibility condition is $\|X_{\cJ^c}^T(y - X\beta^*)\|_{\infty} < \bar{w}_{|\cJ|+1}$ since it can be shown easily that for $\bar{w} = \sum_{j = 1}^N w_j /N $, we have $ \frac{1}{w_1} \|\beta\|_{\infty} \leq \Omega_w^*(\beta) \leq \frac{1}{\bar{w}}\|\beta\|_{\infty}$.

We want to show the OWL dual feasibility is satisfied. Using Theorem 7.5 in \cite{geometric} stated in Appendix with details, we can show that,

$ \|X_{j}^T(y - X\beta^*)\|_{\infty}  \leq \sqrt{32} \log N \frac{\textnormal{aff}(\cS_{\ell},\cS_j)}{\sqrt{d_{\ell}}} \|y - X\beta^*\|_2 $.

We next prove a bound on the size of the residual $\|y-X\beta^*\|$ where $\beta^*$ is defined in Lemma \ref{thm:dual}.

\begin{lemma}\label{lemma:size}\emph{
If $w_1 > 0$, then with probability at least $1 - e^{-\sqrt{N_{\ell}d_{\ell}}}$, we have $\|y - X\beta^*\|_2 \leq c w_1\sqrt{\frac{d_{\ell}}{\log N_{\ell}/d_{\ell}}}.$}
\end{lemma}
Proof details are provided in the Appendix. Lemma \ref{lemma:size} gives with high probability,
\begin{center}
$ \|X_{j}^T(y - X\beta^*)\|_{\infty} \leq w_1 L_N \textnormal{aff}(\cS_{\ell},\cS_j) $
\end{center}
where  $L_N =  c_0 \frac{\log N}{\sqrt{\log \rho_{\ell}}}$, for $j \neq \ell$. Finally using the assumption on the affinity of the subspaces along with union bound and Lemma \ref{thm:dual} completes the proof. 

\subsection{Proof of Theorem \ref{thm:truedisc}}\label{thm2sketch}

The $N_{\ell}$ points on the unit hypersphere can be viewed as forming a $d_{\ell}$-dimensional Random Geometric Graph by placing an edge between points that are $\Delta$-close. To formalize this notion we define a $\Delta$-RGG on the unit hypersphere below (slightly different from the typical RGG defined in the literature \cite{rgg, rgg1}). 
\begin{definition}\emph{
 $\Delta$-Random Geometric Graph (RGG): Place $N_i$ points uniformly at random on the surface of unit hypersphere $\mathbb{S}^{d_i}$. Place an edge between pairs of points (vertices) that are within $\Delta$ Euclidean distance of each other. }
\end{definition}

It can be shown that if enough points are sampled uniformly at random from the unit sphere leads to a fully connected $\Delta$-RGG with high probability. The connectivity and percolation of a typical RGG has been studied (\cite{rgg} and references therein) in the past but the proof of the main result hinges on a fully connected graph defined on the unit hypersphere. Hence, we derive a new bound next.

 \begin{figure}[h]
\centering
\includegraphics[width=0.75\textwidth]{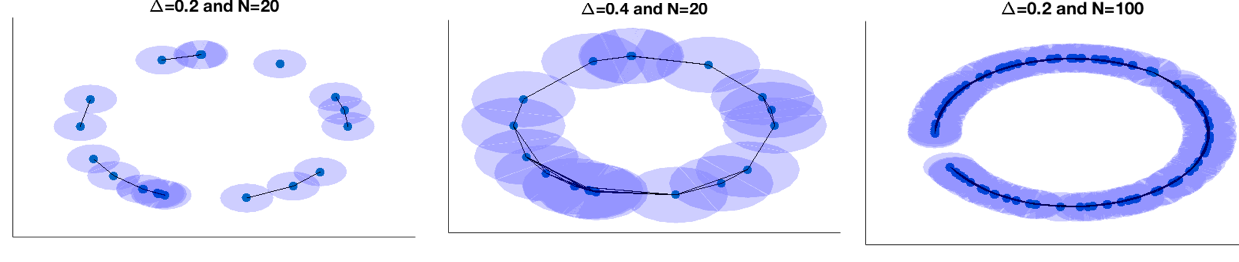}
\caption{Random geometric graphs in $\R^2$ showing effects of different values of $\Delta$ and $N_i$. A $\frac{\Delta}{2}$-hypersphere is drawn centered at every point. An edge is placed between pairs of points if the corresponding hyperspheres intersect. The graph connectivity grows as $\Delta$ or $N_i$ increase.}
\label{fig:rgg}
\end{figure}

 \begin{lemma}\label{thm:rgg}\emph{
If $N_{\ell} > \kappa_1 \Delta^{-d_{\ell}} \log(\Delta^{-d_{\ell}}/\delta)$ points are sampled uniformly at random from the unit hypersphere in $\R^{d_i}$, then the $\Delta$-RGG formed by these points is fully connected with probability at least $1- \delta$.}
\end{lemma}
$\kappa_1$ is a fixed numeric constant. To prove the result we use a covering argument to divide the surface of the unit hypersphere into $m$ equal area patches such that the distance between points in adjacent patches is at most $\Delta$, followed by a multiplicative form of Chernoff's bound to show with high probability at least one point of the uniformly sampled points falls into each patch leading to a fully connected $\Delta$-RGG. Proof details are provided in the Appendix. Using the fact that the Euclidean norm is invariant under multiplication with orthonormal matrix, the distances between points within the subspace can be translated to the ambient subspace. Similarly, we can define a $\Delta$-RGG on the unit hypersphere in $\R^n$. 

We prove a new clustering property for the OWL-Ramp regression next.

\begin{lemma} \label{thm:ramp}\emph{
Let $\wb$ be a solution of the optimization in (\ref{eqn:owlopt}) using OWL-Ramp weights ($\lambda, \Delta> 0$, and $1\leq r \leq N$) and $M := \{j: |\wb_j| =  \max_i |\wb_i|\}$. If elements of $M$ belong to a $\Delta$-connected component on the unit hypersphere in $\R^n$ with cardinality at least $r$, then $|M| \geq r$. }
\end{lemma}
Proof details are provided in the Appendix. This lemma provides a sufficient condition for the largest magnitude cluster in the OWL-Ramp solution to have critical mass. The Euclidean distance condition from Lemma \ref{thm:owl} translates into the $\Delta$-connected component condition in this Lemma. Note that the minimum gap $\Delta_w$ is always smaller than or equal to $\Delta$. It is strictly smaller in most cases. This provides more room for clustering in the OWL solution.   

The proof of Theorem~\ref{thm:truedisc} follows by combining the results from Lemma~\ref{thm:ramp} and Lemma~\ref{thm:rgg} with Theorem~\ref{thm:nofalsedisc} as explained in Appendix.

\section{Numerical results}\label{sec:exp}

\begin{figure}[t!]
    \centering
    \begin{subfigure}[b]{0.44\textwidth}
        \centering
       \includegraphics[width=\textwidth]{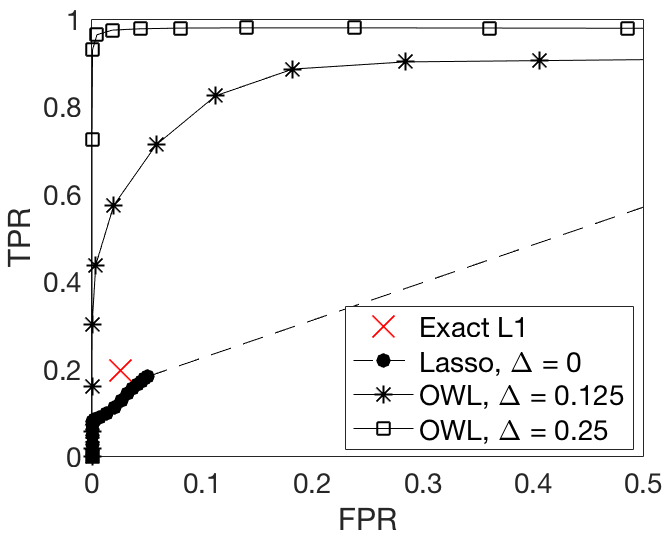}
        \caption{}
    \end{subfigure}%
    ~ 
    \begin{subfigure}[b]{0.44\textwidth}
        \centering
       \includegraphics[width=\textwidth]{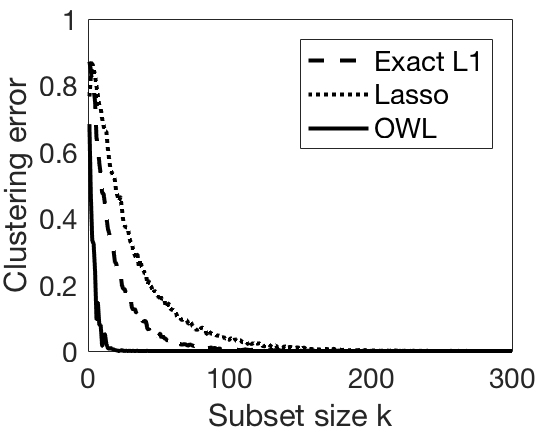}
        \caption{}
        
    \end{subfigure}
    \caption{(a) Trade-off curves for FPR $\leq 0.5$ generated by sweeping through ($\lambda$, $\Delta$) values. Empirical averages of (FPR,TPR) are shown for Exact $\ell_1$, Lasso, and OWL, over $100$ random points. (b) Clustering error for varying $k$ in Algorithm \ref{owlalgo}. The affinity matrix is generated for each method by running only a random subset $k$ of the total $N = 300$ optimizations.}
    \label{fig:rocerror}
\end{figure}

\begin{figure}[t!]
    \centering
    \begin{subfigure}[b]{0.43\textwidth}
        \centering
       \includegraphics[height = 3.3cm]{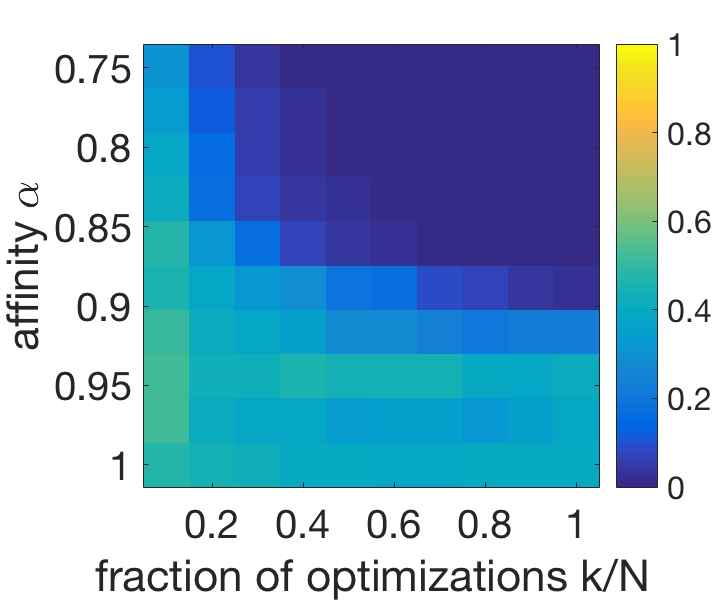}
    \end{subfigure}%
 ~
    \begin{subfigure}[b]{0.47\textwidth}
        \centering
       \includegraphics[height = 3.5cm]{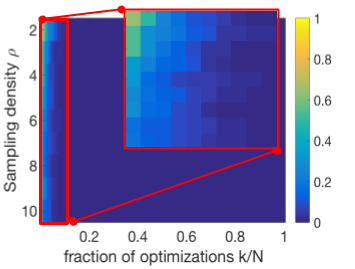}
        
    \end{subfigure}
    \caption{Clustering error for varying $k/N$ in Algorithm \ref{owlalgo} with varying affinity (left) and number of points sampled (right) for OWL regularized subspace clustering.}
    \label{fig:affrho}
\end{figure}

In this section we present numerical results on synthetic data corroborating the theoretical guarantees and providing a better understanding of the behavior of OSC particularly we focus on the OWL-Ramp norm defined in Section~\ref{sec:mainresults}. The subspaces $\cS_1, \cS_2$ and $\cS_3$ are generated with dimension $d = 20$ in $\R^n$ with ambient dimension $n = 40$. 

\textbf{Bases generation method, B1}: The bases $U_1$, $U_2$ and $U_3$ are obtained by choosing, uniformly at random, from the set of all sets of orthonormal vectors in $\R^n$. Since sum of the subspace dimensions exceeds ambient dimension, $n<3d$, this ensures that the subspaces overlap and leads to $\alpha \approx 0.3$.

\textbf{Bases generation method, B2}: To generate subspaces of varying affinity, we follow the method described in \cite{geometric}. The subspaces $\cS_1$, $\cS_2$ and $\cS_3$ are generated using the bases
\[
U_1 
= 
\begin{bmatrix}
I_d\\
\boldsymbol{0}_{d\times d}
\end{bmatrix},
U_2
= 
\begin{bmatrix}
\boldsymbol{0}_{d\times d}\\
I_d
\end{bmatrix},
U_3
= 
\begin{bmatrix}
\textnormal{diag(cos} \boldsymbol{\theta})\\
\textnormal{diag(sin} \boldsymbol{\theta} )
\end{bmatrix},
\]
respectively. Where the principal angles are set in such a way that the normalized affinity decreases linearly from 1 to 0.75 and diag(cos $\boldsymbol{\theta}$) corresponds to a diagonal matrix with diagonal entries equal to cos $\theta_i$.

We select $\rho = 5$ points per subspace dimension (unless stated otherwise), \textit{i.e,} $N_i = \rho d = 100$ points uniformly at random from each subspace using the respective bases. 

\textbf{Choice of hyper-parameters.} See Section~\ref{sec.tuningparameters} for discussion on the choice of tuning parameters $\Delta$, $\lambda$ and $r$ guided by the theory in this paper and past literature. We fix the ramp parameter, $r = N/3 = 100$, in the experiments.

\textbf{The FPR-TPR trade off.} In order to compare the performance of the optimizations in (\ref{eqn:l1opt}), (\ref{eqn:lasso}) and (\ref{eqn:owlopt}), we generate $N$ columns from subspaces generated using B1 to run $N$ optimizations of each method. For each data point, we sweep through different values of the tuning parameters ($\lambda, \Delta$). For $\lambda = 0$, all points are selected and as $\lambda$ is increased fewer points are given non-zero weight. Let $\wb$ denote the solution to one of the optimizations. We plot the empirical averages of False Positive Rate (FPR)  $= \|\wb_{S^c}\|_0/|S^c|$ and True Positive Rate (TPR) $= \|\wb_S\|_0/|S|$ where $\wb_S$ is the part of $\wb$ supported on indices of the points from the same subspace and $\wb_{S^c}$ is supported on its complement. A non-zero entry in $\wb_{S}$ is a true discovery and likewise a non-zero entry in $\wb_{S^c}$ is a false discovery. By definition $|S| = N_i$ and $|S^c| = N - N_i$.

Figure \ref{fig:rocerror} (a) shows the Receiver Operating Characteristic (ROC) curve plotting TPR versus FPR. The solution of an exact $\ell_1$ minimization is shown as one point on the curve since the sparsity of the solution cannot be changed by varying tuning parameters. Note for any $\lambda>0$, the lasso solution will include {\em at most} $n = 40$ non-zero entries, since the number of selected columns will not exceed the ambient dimension. If $\lambda=0$, then lasso will select all columns. Thus, lasso curve beyond $d$ selections is shown as a dashed line, which extends linearly to the point (FPR,TPR) = $(1,1)$.  As suggested by theory and demonstrated by the plots, for a fixed false positive rate, OWL can achieve a much higher ($\boldsymbol{9\times}$) true positive rate than Lasso.

\textbf{Effect of the size of the optimization subset $k$.} To observe the effect of the size of the optimization subset $k$, we look at the clustering error produced by running a subset $k$ of the total $N$ optimizations for each method for different values of $k$. The subspaces are generated using B1. We take the symmetrized affinity matrix, $W$, generated by each method, assuming knowledge of number of subspaces, apply a spectral clustering method to obtain the clusters. 
The clustering error is measured as the fraction of misclassified points from the total number of points. Figure \ref{fig:rocerror} (b) shows how the clustering error varies with $k$ for SSC with Exact $\ell_1$, Lasso, and OWL regularized sparse linear regression steps.  We report the empirical average of the clustering error over 100 random choices of subsets. By virtue of the clustering property of OWL, we see that the clustering error is low for subset sizes of the order of the number of subspaces. As demonstrated in Figure \ref{fig:rocerror} (b), to achieve a clustering error of less than 0.01, Lasso requires $\boldsymbol{8}$ times more optimizations than OWL. The plots suggest that the OWL approach could potentially achieve a speedup of up to $\boldsymbol{20\times}$.

\begin{figure}[t!]
\centering
\includegraphics[width=0.75\textwidth]{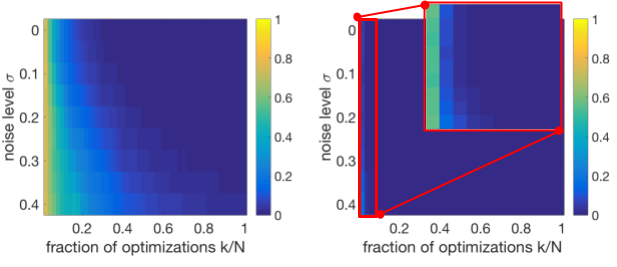}
\caption{Clustering error for varying $k$ in Algorithm \ref{owlalgo} versus varying noise levels, $\sigma$, for Lasso (left) and OWL (right) minimization based SSC. Affinity matrix for each method is generated by running only a random subset $k$ of total $N = 300$ optimizations.}
\label{fig:noise}
\end{figure}

\textbf{Effect of affinity and number of points sampled.} We vary the amount of correlation between subspaces and the number of points sampled from each subspace and study its effect on the clustering error for different values of $k$ in Figure \ref{fig:affrho}. The subspaces with varying affinity are generated using the method described in B2 and subspaces for varying $\rho$ are generated using B1. The affinity is varied in the range $\alpha \in [0.75, 1]$ where $\alpha = \max_{\ell} \alpha_{\ell}$ and we vary the sampling density $\rho \in [2,10]$. Recall that the affinity is low when the subspaces are nearly orthogonal and high when the subspaces overlap significantly (it is equal to one when one subspace is contained in the other). The tuning parameters are fixed throughout the experiment. As expected, the clustering error increases for higher values of affinity and OWL ($\Delta = 0.01$) provides no benefit over Lasso ($\Delta = 0$). On the other hand, OWL produces small values of error for most values of $\rho$. 
\begin{table}[t!]
\small
\caption{Clustering error (\%) of different algorithms on the Hopkins 155 dataset.}
\centering
\begin{tabular}{ |c|c|c|c|c|c|c| } 
 \hline
 L &  Algorithm & SSC  & OMP & TSC & NSN & OSC \\ 
 \hline
 \multirow{2}{*}{2} &  Mean & 1.52   &  16.92 & 18.44 & 3.62 & 1.39\\ 
 & Median & 0.00 &  12.77  & 16.92  & 0.00 & 0.00 \\
 \hline
 \multirow{2}{*}{3} &  Mean &  4.40  & 27.96 & 28.58 & 8.28 & 5.21 \\ 
& Median & 0.56 & 30.98 &29.67  & 2.76 & 1.02 \\
 \hline
\end{tabular}
\label{table:1}
\end{table}
\begin{table}[t!]
\small
\caption{Mean clustering error (\%) on the Hopkins 155 dataset with the 4$L$-dimensional data points obtained by applying PCA.}
\centering
\begin{tabular}{ |c|c|c| } 
 \hline
 Motions (L) & SSC & OSC \\ \hline
   2  & 1.83  & 1.49 \\ 
 3 & 4.40  & 5.16 \\ 
All  & 2.41  & 2.32 \\
 \hline
\end{tabular}
\label{table:2}
\end{table}

\begin{figure}[t!]
    \centering
    \begin{subfigure}[b]{0.44\textwidth}
        \centering
       \includegraphics[width=0.99\textwidth]{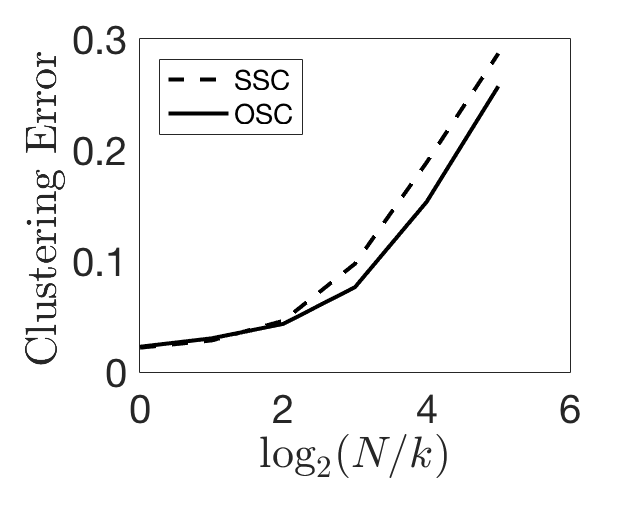}
        \caption{}
    \end{subfigure}%
    ~ 
    \begin{subfigure}[b]{0.44\textwidth}
        \centering
       \includegraphics[width=0.99\textwidth]{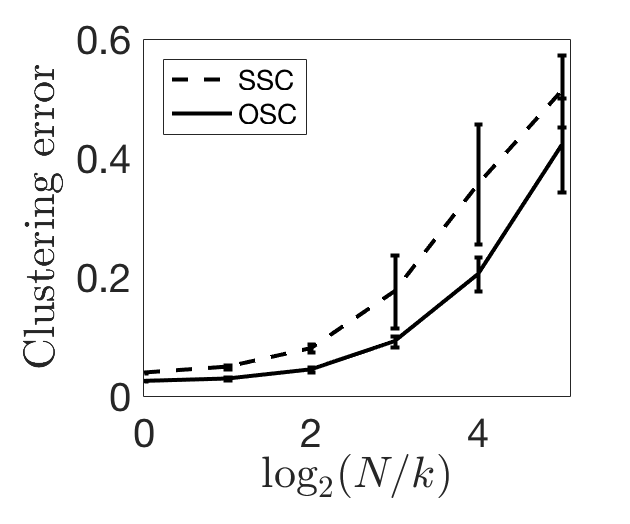}
        \caption{}
        
    \end{subfigure}
    \caption{(a) Mean clustering error for Hopkins dataset.  (b) Mean clustering error for MNIST dataset.}
    \label{fig:hopkins}
\end{figure}

\textbf{Effect of noise.} To demonstrate the performance of the algorithms under noise, we perturb each unit norm data point $x_i$ by a noisy vector chosen uniformly at random from the sphere of radius $\sigma$ and normalize again to have unit norm. The noisy model has been studied in \cite{noisy} using the Lasso version of SSC. Figure \ref{fig:noise} shows the empirical average of clustering errors obtained by the three methods across different subspaces generated using B1 by varying the noise level and the number of optimizations run. The OWL method with $k = 10$ achieves low clustering error for noise levels as high as $\sigma = 0.4$ providing a speedup of up to $\boldsymbol{30\times}$.

\section{Experiments with Real Data}\label{sec:realexp}

We compare our algorithm with the existing ones in the applications of motion segmentation and clustering handwritten digits. 

For motion segmentation, we used Hopkins155 dataset \cite{tron2007benchmark}. It contains 155 video sequences of 2 or 3 motions. Table \ref{table:1} summarizes clustering errors of different algorithms on the Hopkins155 dataset. For ease of comparison, the error values for TSC, SSC-OMP and NSN are populated from experiments conducted in \cite{park2014greedy} where the authors optimized the parameters for the existing algorithms. The parameters for SSC were set as provided in the source code. The OWL parameters are set as follows: for all real data experiments we use the same $\lambda$ value as SSC, we set $r = N/4$ (rounded off to nearest integer) and set $\Delta$ such that $w_1 = \lambda + r \Delta = 2\lambda$ or $4\lambda$, whichever, if any, leads to a non-trivial solution.
Figure \ref{fig:hopkins}(a) shows the effect of running a subset of $k$ optimizations selected at random from the set of all $N$ optimizations for $k \in \{N, N/2, N/4, \dots, N/32\}$. OSC outperforms or performs about the same as SSC in most cases with comparable running times. The SSC and OSC parameters in this application lead to relatively dense affinity matrices in both cases. This results in low clustering errors with fewer optimizations supporting the theory.
Table \ref{table:2} summarizes results of applying the SSC and OSC algorithms to the dataset projected into a 4$L$-dimensional subspace using PCA suggesting further speedups without loss in performance.

We also use the MNIST test data set \cite{mnist} that contains 10,000 centered 28 $\times$ 28 pixel images of handwritten digits, \textit{i.e.,} it contains many more points per subspace. The empirical mean and standard deviation of the CE are computed by averaging over 100 of the following problem instances. We choose the digits \{2, 4, 8\} and we choose $k$ optimizations uniformly at random from the set of all optimizations for $k \in \{N, N/2, N/4, \dots, N/32\}$. Here we use the default parameters for SSC and set OWL parameters in similar fashion to the motion segmentation experiments. The results are summarized in Figure \ref{fig:hopkins}(b).


\section{Conclusion}
In this paper we have proposed a new approach to subspace
  clustering based on Ordered Weighted $\ell_1$ (OWL) minimization which performs
  simultaneous regression and clustering of correlated variables. The clustering property of the OWL, combined with ideas from random geometric graph theory, allows us to prove that OSC tends to select more points from the correct subspaces in each regression compared to SSC. In the ideal case, where $L$ subspaces are orthogonal and 
the number of points per subspace is sufficiently large, then OSC can succeed with just $L \ll N$ optimizations (gain factor of roughly $N/L$). This key feature of OSC makes accurate clustering possible based on regression solutions for only a small subset of the total dataset, significantly reducing the computational complexity
  compared to SSC.

\section*{Acknowledgments}
The authors thank Ulaz Ayaz for fruitful discussions.

\bibliographystyle{abbrv}
\bibliography{root}

\appendix
\section*{Appendix}

\section{Properties of OWL}

Subgradient condition for OWL:
$$ z_i = \textnormal{sign}(\beta_i) (P_\beta w)_i, \textnormal{ if } \beta_i \neq 0, \textnormal{ and } z_i = 0, \textnormal{otherwise.}$$
When defined as above, we can write the subdifferential vector as
$$ z = (P_\beta w) \cdot \textnormal{sign}(\beta) $$

\section{Proof of Lemma \ref{thm:dual}}

\begin{proof}

Let $\cT = \{i | \beta_i^* \neq 0 \}$. Note that $\cT \subseteq \cJ$.\\

From optimality conditions, we have $ X_\cJ^T(y - X\beta^*) = (P_{\beta^*}w)_\cJ \cdot \textnormal{sign}(\beta^*_\cJ)$.\\

Consider a perturbation $\beta^* + th$. \\
Let $P_{\beta^*}$ and $P_{\beta^* + th}$ be permutation matrices such that  $\Omega_w(\beta^*) = (P_{\beta^*}w)^T |\beta^*| $ and $\Omega_w(\beta^* + th) = (P_{\beta^* + th}w)^T |\beta^* +th| $. Notice that these permutation matrices may not be unique. 

For $t > 0$ sufficiently small such that $ \textnormal{sign}(\beta^*_{\cT}) =  \textnormal{sign}(\beta^*_{\cT} + th_{\cT})$ and the group ordering doesn't change i.e, there exist $P_{\beta^*}$ and $P_{\beta^* + th}$ such that $(P_{\beta^*}w)_{\cT}  = (P_{\beta^* + th}w)_{\cT} $,
\begin{align*}
&\Omega_w(\beta^* + th) \\
& = \langle (P_{\beta^* + th}w)\cdot \textnormal{ sign }(\beta^* +th),  \beta^* +th \rangle \\
& = \langle (P_{\beta^* + th}w)_{\cT} \cdot \textnormal{sign}(\beta_{\cT}^* + th_{\cT}), \beta_{\cT}^* + th_{\cT} \rangle  + t (P_{\beta^* + th}w)_{\cT^c}^T |h_{\cT^c}|\\
& = \langle (P_{\beta^* + th}w)_{\cT} \cdot \textnormal{sign}(\beta_{\cT}^*), \beta_{\cT}^* + th_{\cT} \rangle  + t (P_{\beta^* + th}w)_{\cT^c}^T |h_{\cT^c}|\\
&= \Omega_w(\beta^*) + t\langle (P_{\beta^*}w)_{\cT} \cdot \textnormal{sign}(\beta^*_{\cT}), h_{\cT} \rangle + t (P_{\beta^* + th}w)_{\cT^c}^T |h_{\cT^c}|\\
&= \Omega_w(\beta^*) + t\langle (P_{\beta^*}w)_\cJ \cdot \textnormal{sign}(\beta^*_\cJ), h_\cJ \rangle + t\Omega_{(P_{\beta^* + th}w)_{\cT^c \downarrow}}(h_{\cT^c})\\
&\geq \Omega_w(\beta^*) + t\langle (P_{\beta^*}w)_\cJ \cdot \textnormal{sign}(\beta^*_\cJ), h_\cJ \rangle + t\Omega_{w'}(h_{\cJ^c})
\end{align*}
where the last inequality follows since $\cJ^c \subseteq \cT^c$ and $ (P_{\beta^* + th}w)_{\cT^c\downarrow}  = [w_{|\cT| + 1}, \dots, w_{|\cJ|}, w_{|\cJ| + 1},\dots, w_N]$. The weights on $h_{\cJ^c}$ will only decrease. Recall from optimality conditions we have $ X_{\cJ}^T(y - X\beta^*) = (P_{\beta^*}w)_{\cJ} \cdot \textnormal{sign}(\beta^*_{\cJ})$, this gives us

\begin{align*}
 &\frac{1}{2} \|y - X(\beta^* + th)\|_2^2 + \Omega_w(\beta^* + th) \\
 & = \frac{1}{2} \|y - X\beta^*\|^2_2 + \frac{t^2}{2}\|h\|_2^2 - t \langle X^T(y - X\beta^*), h \rangle + \Omega_w(\beta^* + th)\\
&\geq \frac{1}{2} \|y - X\beta^*\|^2_2 + \Omega_w(\beta^*)  + \frac{t^2}{2}\|h\|_2^2  + t\Omega_{w'}(h_{\cJ^c}) - \langle X_{\cJ^c}^T(y - X\beta^*), h_{\cJ^c} \rangle
\end{align*}

Let $X_{\cJ^c}^T(y - X\beta^*) =: \epsilon_{\cJ^c}$ and if $h_{\cJ^c} \neq 0$. Consider the last term,

\begin{align*}
\Omega_{w'}(h_{\cJ^c}) - \langle \epsilon_{\cJ^c}, h_{\cJ^c} \rangle & \geq \Omega_{w'}(h_{\cJ^c}) -  \Omega^*_{w'}(\epsilon_{\cJ^c}) \Omega_{w'}(h_{\cJ^c}) \\
&> 0
\end{align*}

from assumption $\Omega^*_{w'}(\epsilon_{\cJ^c}) < 1$.\\

This implies $ \frac{1}{2} \|y - X(\beta^* + th)\|_2^2 + \Omega_w(\beta^* + th) > \frac{1}{2} \|y - X\beta^*\|^2_2 + \Omega_w(\beta^*) $ and the claim follows.
 
\end{proof}

\section{Proof of Lemma \ref{lemma:size}}

\begin{proof}
Define:
$$ \wb^{(1)} = \textnormal{arg}\min_{\beta^{(1)}} \frac{1}{2} \|y - X^{(1)} \beta^{(1)}\|_2^2 + \Omega_w(\beta^{(1)}) $$
$$ \xb^{(1)} = \textnormal{arg}\min_{\beta^{(1)}} \Omega_w(\beta^{(1)}) \textnormal{ s.t. } y =   X^{(1)} \beta^{(1)}$$

Since $\wb^{(1)}$ minimizes the first optimization and $y =   X^{(1)} \xb^{(1)}$, we have the following
$$ \frac{1}{2} \|y - X^{(1)} \wb^{(1)}\|_2^2 + \Omega_w(\wb^{(1)}) \ \leq \ \frac{1}{2} \|y - X^{(1)} \xb^{(1)}\|_2^2 + \Omega_w(\xb^{(1)}) \ =  \ \Omega_w(\xb^{(1)})$$

Let $ h = \wb^{(1)} - \xb^{(1)}$, then 
$$ \frac{1}{2} \|y - X^{(1)} \wb^{(1)}\|_2^2 \ = \ \frac{1}{2} \| X^{(1)} h \|_2^2 \  \leq \ \Omega_w(\xb^{(1)}) - \Omega_w(\xb^{(1)} + h) $$

Let $P$ and $Q$ be permutation matrices such that:
$$\Omega_w(\wb^{(1)}) = \langle P_{\wb^{(1)}}w,  |\wb^{(1)} |\rangle$$ 
$$\Omega_w(\xb^{(1)}) = \langle P_{\xb^{(1)}}w,  |\xb^{(1)}| \rangle$$ 

Note that by definition of OWL norm $\langle P_{\xb^{(1)}}w,  |\wb^{(1)}| \rangle \leq \langle P_{\wb^{(1)}}w, | \wb^{(1)}| \rangle$. Let $S$ be the support of $\xb$, then
\begin{align*}
\Omega_w(\xb^{(1)} + h) -  \Omega_w(\xb^{(1)}) & = \langle P_{\wb^{(1)}}w,  |\xb^{(1)} + h| \rangle - \langle P_{\xb^{(1)}}w, | \xb^{(1)}| \rangle\\
& \geq \langle P_{\xb^{(1)}}w, |\xb^{(1)} + h| - | \xb^{(1)}| \rangle\\
& = \langle (P_{\xb^{(1)}}w)_S, |\xb^{(1)}_S + h_S| - | \xb^{(1)}_S| \rangle + \langle (P_{\xb^{(1)}}w)_{S^c}, | h_{S^c}| \rangle\\
& \geq  \langle \textnormal{sign}( \xb^{(1)}_S)\cdot(P_{\xb^{(1)}}w)_S,  h_S \rangle + \langle (P_{\xb^{(1)}}w)_{S^c}, | h_{S^c}| \rangle 
\end{align*}
 
Plugging into the inequality above gives

$$ \frac{1}{2} \| X^{(1)} h \|_2^2  \leq -  \langle \textnormal{sign}( \xb^{(1)}_S)\cdot(P_{\xb^{(1)}}w)_S,  h_S \rangle - \langle (P_{\xb^{(1)}}w)_{S^c}, | h_{S^c}| \rangle$$

since $\xb^{(1)}$ is optimal there exists $\nu$ such that 
$$ v = X^{(1)^T}\nu, v_S = \textnormal{sign}( \xb^{(1)}_S)\cdot(P_{\xb^{(1)}}w)_S \textnormal{ and } \Omega^*_{(P_{\xb^{(1)}}w)_{S^c}}(v_{S^c}) \leq 1 $$

using this 
$$ \langle \textnormal{sign}( \xb^{(1)}_S)\cdot(P_{\xb^{(1)}}w)_S,  h_S \rangle = \langle v_S,  h_S \rangle = \langle \nu, X^{(1)}h \rangle- \langle v_{S^c},  h_{S^c} \rangle $$

Thus we have
\begin{align*}
 |\langle \textnormal{sign}( \xb^{(1)}_S)\cdot(P_{\xb^{(1)}}w)_S,  h_S \rangle| &\leq |\langle \nu, X^{(1)}h \rangle| + |\langle v_{S^c},  h_{S^c} \rangle| \\
 & \leq \|X^{(1)}h\|_2\|\nu\|_2 + \Omega^*_{(P_{\xb^{(1)}}w)_{S^c}}(v_{S^c}) \Omega_{(P_{\xb^{(1)}}w)_{S^c}}(h_{S^c}) \\
 & \leq  \|X^{(1)}h\|_2\|\nu\|_2 + \Omega_{(P_{\xb^{(1)}}w)_{S^c}}(h_{S^c}) 
 \end{align*}
Plugging back into 
$$ \frac{1}{2} \| X^{(1)} h \|_2^2  \leq \|X^{(1)}h\|_2\|\nu\|_2 + \Omega_{(P_{\xb^{(1)}}w)_{S^c}}(h_{S^c}) - \langle (P_{\xb^{(1)}}w)_{S^c}, | h_{S^c}| \rangle$$

Note that since $\xb_{S^c} = 0$, we have $h_{S^c} = \wb_{S^c}$ and $ \Omega_{(P_{\xb^{(1)}}w)_{S^c}}(h_{S^c}) = \langle (P_{\xb^{(1)}}w)_{S^c}, | h_{S^c}| \rangle $. Thus it follows that

$$ \frac{1}{4} \| X^{(1)} h \|_2^2  \leq \|\nu\|_2^2 \leq c w_1^2 \frac{d}{\log(\frac{N}{d})} $$

where proof Last inequality is shown next.

\end{proof}

Consider the exact OWL norm minimization problem,
$$ \min_\beta \ \Omega_w(\beta) \ \textnormal{ s.t. } \ y = X\beta $$
and its dual 
$$ \max_{\nu} \ \langle y, \nu \rangle \textnormal{ s.t. } \ \Omega^*_w(X^T \nu )  \ \leq 1 $$

Any dual feasible point $\nu$ satisfies,
$$\frac{1}{w_1}\|X^T\nu\|_{\infty} \leq \Omega^*_w(X^T \nu )  \ \leq 1$$

Thus, $\frac{1}{w_1}\nu \in K^o$ which implies 
$$\|\frac{1}{w_1}\nu\|_2 \leq R(K^o) = \frac{1}{r(K)}$$
where $K = $conv({$x_i$), $r(K)$ is its inradius, $K^o$ is the polar set and $R(K^o)$ is its circumradius. Equality follows since $R(K^o).r(K) = 1$.

We get $$ \|\nu\|_2 \leq \frac{w_1}{r(K)}$$

finally, using standard results (like Lemma 7.4 in \cite{geometric}) we get with probability at least $1- e^{-\sqrt{Nd}}$,

$$ \|\nu\|_2^2 \leq c w_1^2 \frac{d}{\log(\frac{N}{d})} $$
where c is a constant.

\section{Proof of Lemma \ref{thm:rgg}}

\begin{proof}

Suppose we divide the surface of the unit hypersphere into $m$ patches of equal surface area. Then the probability that a uniformly sampled point falls into a particular patch is $p = 1/m$.

Let $C =$ the number of times a sample falls in a particular patch in $N_{\ell}$ independent trials. The multiplicative form of Chernoff's bound is 

        $$P(C \leq (1-b) \mu)  \leq \exp\left(-\frac{b^2 \mu}{2}\right),  \textnormal{ for any } 0 < b \leq 1$$

where $\mu = E[C] = N_{\ell} p = N_{\ell}/m$.  Taking $b=1$ we get

        $$P(C \leq 0)  \leq \exp\left(-\frac{N_{\ell}}{2m}\right)$$

Let $\delta' \geq \exp\left(-\frac{N_{\ell}}{2m}\right)$.  It follows that if $N_{\ell} \geq 2m \log(1/\delta')$, then with probability at least $1-\delta'$ there is at least one sample falling in the patch.  Now if we want this to hold for all $m$ patches, we can union bound to get the following.

$$ P(\cup_{i=1}^m C_i \leq 0 ) \leq \sum_{i=1}^m P(C_i \leq 0 ) \leq m \delta' =: \delta $$

If $N_{\ell} \geq 2m \log(m/\delta)$, then with probability at least $1-\delta$ there is at least one sample in each of the $m$ patches. 

If we construct the patches such that the distance between any point in a certain patch and any point in adjacent patches is always less than or equal to $\Delta$, this gives us a completely connected $\Delta$-RGG.

\textbf{Construction of the patches}: $\epsilon$-covering number ($N^{\epsilon}_{d_{\ell}}(\mathbb{S}^{d_{\ell}})$) of unit sphere in $\R^{d_{\ell}}$ has the following property:

$$N^{\epsilon}_{d_{\ell}}(\mathbb{S}^{d_{\ell}}) \leq 3\epsilon^{-d_{\ell}}$$

so we need at least $m = O(\Delta^{-d_{\ell}})$ patches for $\epsilon < \Delta/2$ and $N_{\ell}  > \kappa_1 \Delta^{-d_{\ell}} \log( \Delta^{-d_{\ell}}/\delta)$ points in the subspace to say with probability at least $1-\delta$, there will be a point in every patch on the surface of the unit sphere leading to a connected $\Delta$-RGG.
\end{proof}

\section{Proof of Lemma \ref{thm:ramp}}

\begin{proof}

\begin{description}
\item 

By contradiction, assume $|M| < r$.

Let $r' = r - |M|$ and let $G$ be the $\Delta$-connected component containing elements of $M$. 

Define $G_0 := M$. For $m = 1, \dots, r'$, define 
$$ G_m = G_{m-1} \cup j \textnormal{ for some } j\in G, j \notin G_{m-1},  \exists i \in G_{m-1} \textnormal{ such that }  \| x_i - x_ j\| < \Delta.$$

\item By definition, we have for $m = 0, \dots,  r'$ that $G_m \subseteq G$.

\item If for some $m \in [r']$, the set $G_{m-1} = M = \textnormal{arg} \max_i |\wb_i| $, then by definition there exists $j \in G_m$ such that $|\wb_j| < |\wb_i|$, $\forall i \in G_{m-1}$.

\item In the rest of the proof, we show a contradiction arises and completes the proof by induction.

\end{description}

Consider the following alternative solution, $\tb \in \R^p$, such that
\begin{align*}
\tb_j& =  \wb_j + \epsilon,\\
\tb_i &= \wb_i - \epsilon, \textnormal{ for the } i \in G_{m-1} \textnormal{ that is $\Delta$ close to } j\\
\tb_k &= \wb_k \textnormal{ for } k \neq i,j
\end{align*}

where $\epsilon \in (0, \min\{\frac{|\wb_i| - |\wb_j|}{2}, |\wb_i| - |\widehat{\mu}_{2}|\})$ where $|\widehat{\mu}_{2}|$ is the second largest unique magnitude of $\wb$, this ensures the components of $M$ stay in the top $r$ in the alternative solution.

From Lemma 2.3 of \cite{aistatsOWL} we have
\begin{align*} 
L(\tb) - L(\wb) & =  \|y - X \tb\|_2^2 -  \|y - X \wb\|_2^2\\
& \leq   \epsilon \|y\| \| x_i - x_j \|\\
& < \epsilon \Delta 
\end{align*}
where the last inequality follows by assumption $\| x_i - x_j \| < \Delta$ and $\|y\| = 1$.

Also, we have 
\begin{align*}
 \Omega_w(\tb) - \Omega_w(\wb) &= \lambda \|\tb\|_1 + \Delta \sum_{i=1}^{d} (d-i+1) |\tb_{[i]}| - \lambda \|\wb\|_1 - \Delta \sum_{i=1}^{d} (r-i+1) |\wb_{[i]}|\\
 & = \Delta \sum_{i=1}^{r} (r-i+1) |\tb_{[i]}|- \Delta \sum_{i=1}^{r} (r-i+1) |\wb_{[i]}|\\
& \leq - \Delta \epsilon 
\end{align*} 
where the second last equality follows since $|\wb_i|$ is in the top $r$ magnitudes of $\wb$ and the definition of $\epsilon$ ensures $|\tb_i|$ is in the top $r$ magnitudes of $\tb$ along with a variant of Lemma 2.1 when $\textnormal{sign}(\wb_i) = \textnormal{sign}(\wb_j)$ or a variant of Lemma 2.2 when $\textnormal{sign}(\wb_i) \neq \textnormal{sign}(\wb_j)$  from \cite{aistatsOWL}  (using the fact that $w_{\ell+a} - w_{m-b} \geq \Delta $ and $w_{\ell+a} - w_{m+b} \geq \Delta $  since we ensure that $\ell+a > r$ and $m-b \geq r$.) 

Putting these together we have 

$$ L(\tb) - L(\wb) +  \Omega_w(\tb) - \Omega_w(\wb) < 0$$

This contradicts our assumption that $\wb$ is the minimizer of  $L(\beta) + \Omega_w(\beta)$.
\end{proof}

\section{Proof details of Theorem \ref{thm:nofalsedisc} continued}

We start with the deterministic lemma stated in Lemma \ref{thm:dual} that introduces the OWL dual feasibility condition. 

The lemma tells us that if the OWL dual feasibility condition $\Omega^*_{w'}(X_{\cJ^c}^T(y - X\beta^*)) < 1$ is satisfied for $\cJ = \{j: X_j \in \cS_{\ell}\}$, then we have no false discoveries. To prove Theorem \ref{thm:nofalsedisc}, it suffices to show that the dual feasibility condition is satisfied. One sufficient condition to satisfy the OWL dual feasibility condition is $\|X_{\cJ^c}^T(y - X\beta^*)\|_{\infty} < \bar{w}_{|\cJ|+1}$ since it can be shown easily that for $\bar{w} = \sum_{j = 1}^N w_j /N $, we have
$$ \frac{1}{w_1} \|\beta\|_{\infty} \leq \Omega_w^*(\beta) \leq \frac{1}{\bar{w}}\|\beta\|_{\infty}$$

To show the dual feasibility is satisfied we use the following result. 

\begin{lemma}[Theorem 7.5 in \cite{geometric}]
Let $A \in \R^{d_{\ell} \times N_{\ell}}$ be a matrix with columns sampled uniformly at random from the unit sphere of $\R^{d_{\ell}}$, $v \in \R^{d_j}$ be a vector sampled uniformly at random from the unit sphere of $\R^{d_j}$ and independent of $A$ and $\Sigma \in \R^{d_{\ell} \times d_j}$ be a deterministic matrix. We have 
$$\| A^T \Sigma v\|_{\infty} \leq \sqrt{\log a \log b} \frac{\|\Sigma\|_F}{\sqrt{d_{\ell}}\sqrt{d_j}}, $$
with probability at least $1 - \frac{2}{\sqrt{a}} - \frac{2N_{\ell}}{\sqrt{b}}$.

\end{lemma}

We can use this as follows. Suppose $\Sigma = U^{{(j)}^T} U^{({\ell})}$, where $U^{(j)}$ is an orthogonal basis for $\cS_j$ and $U^{(\ell)}$ for $\cS_{\ell}$ respectively. By definition, $\|\Sigma\|_F = \sqrt{d_{\ell} \wedge d_j} \textnormal{aff}(\cS_j,\cS_{\ell})$. Consider 
$$ \|X_{j}^T(y - X\beta^*)\|_{\infty}  = \| A^T \Sigma v\|_{\infty} \|y - X\beta^*\|_2 $$

Using the Lemma with $a = N^4, b = N^8$, we have with probability at least $1 - 4/N^2$
$$ \|X_{j}^T(y - X\beta^*)\|_{\infty}  \leq \sqrt{32} \log N \frac{\textnormal{aff}(\cS_{\ell},\cS_j)}{\sqrt{d_{\ell}}} \|y - X\beta^*\|_2 $$

Lemma \ref{lemma:size} gives, with probability at least $1- e^{-\sqrt{N_{\ell} d_{\ell}}} - 4/N^2$,
$$ \|X_{j}^T(y - X\beta^*)\|_{\infty} \leq w_1 L_N \textnormal{aff}(\cS_{\ell},\cS_j) $$
where  $L_N =  c_0 \frac{\log N}{\sqrt{\log \rho_{\ell}}}$, for $j \neq \ell$.

Finally using the assumption on the affinity of the subspaces we get $ \|X_{j}^T(y - X\beta^*)\|_{\infty} \leq \bar{w}_{|\cJ|+1}$ with probability at least $1- e^{-\sqrt{N_{\ell} d_{\ell}}} - 4/N^2$ which along with union bound and Lemma \ref{thm:dual} completes the proof of Theorem \ref{thm:nofalsedisc}. \\

\section{Proof details of Theorem \ref{thm:truedisc} continued}\label{thm2sketch}

Let $\cJ$ be set of indices of columns belonging to the subspace, $\cS_{\ell}$. From Theorem \ref{thm:nofalsedisc}, we have with high probability the coefficients of $\wb_{\cJ^c} = 0$.  Lemma \ref{thm:rgg} with the assumption on the number of points sampled gives us that the $\Delta$-RGG formed by the points $X_\cJ$ on the unit hypersphere is fully connected. The connected component has cardinality at least $N_{\ell}$. If $\max_i |\wb_i| = 0$, then the claim is trivial so suppose $\max_i |\wb_i| > 0$, then $M \subseteq \cJ$ and the claim follows from $r \leq N_{\ell}$ and Lemma \ref{thm:ramp}.

\end{document}